\pdfoutput=1

\documentclass[10pt,twocolumn,letterpaper]{article}

\usepackage[pagenumbers]{wacv}

\usepackage{graphicx}
\usepackage{amsmath}
\usepackage{amssymb}
\usepackage{amsthm}
\usepackage{booktabs}
\usepackage{multirow}
\usepackage[protrusion=true,expansion=true,stretch=10,shrink=10]{microtype}

\definecolor{wacvblue}{rgb}{0.21,0.49,0.74}
\usepackage[pagebackref,breaklinks,colorlinks,allcolors=wacvblue]{hyperref}

\usepackage[capitalize]{cleveref}
\crefname{section}{Sec.}{Secs.}
\Crefname{section}{Section}{Sections}
\Crefname{table}{Table}{Tables}
\crefname{table}{Tab.}{Tabs.}

\crefname{proposition}{Prop.}{Props.}
\Crefname{proposition}{Proposition}{Propositions}

\def\wacvPaperID{2372}
\def\confName{WACV}
\def\confYear{2027}

\AddToShipoutPicture*{\AtPageUpperLeft{\put(\LenToUnit{0.75in},\LenToUnit{-0.62in}){\parbox[t]{7in}{\centering\fontsize{6.5}{7.5}\selectfont
Accepted version of a paper to appear in the 2027 IEEE/CVF Winter Conference on Applications of Computer Vision (WACV). \textcopyright~2027 IEEE. Personal use of this material is permitted. Permission from IEEE must be obtained for all other uses, in any current or future media, including reprinting/republishing this material for advertising or promotional purposes, creating new collective works, for resale or redistribution to servers or lists, or reuse of any copyrighted component of this work in other works.}}}}

\begin{document}

\title{A Universal Raw-IQ Radar Encoder for Cross-Sensor Human Sensing}

\author{
Jonas Leo Mueller\textsuperscript{1,3,5} \hspace{0.4em}
Dario Zanca\textsuperscript{3} \hspace{0.4em}
Sebastian Hoefler\textsuperscript{3,5} \hspace{0.4em}
Jan Petermann\textsuperscript{1,5} \hspace{0.4em}
Thomas Altstidl\textsuperscript{2,3,5} \\[2pt]
Naga Venkata Sai Jitin Jami\textsuperscript{3,5} \hspace{0.4em}
Emmanuelle Salin\textsuperscript{2,5} \hspace{0.4em}
Bjoern M. Eskofier\textsuperscript{1,2,3,4,5}
\\[0.5em]
{\footnotesize \textsuperscript{1}Institute for AI in Medicine, LMU Hospital, Munich, Germany}\\
{\footnotesize \textsuperscript{2}Chair of AI-supported Therapy Decisions, Ludwig-Maximilians-Universit\"at M\"unchen, Munich, Germany}\\
{\footnotesize \textsuperscript{3}Department Artificial Intelligence in Biomedical Engineering, Friedrich-Alexander-Universit\"at Erlangen-N\"urnberg (FAU), Erlangen, Germany}\\
{\footnotesize \textsuperscript{4}Institute of AI for Health, Helmholtz Zentrum M\"unchen, Neuherberg, Germany \enspace \textsuperscript{5}Munich Center for Machine Learning (MCML), Munich, Germany}\\
{\footnotesize \href{mailto:jonas.leo.mueller@fau.de}{jonas.leo.mueller@fau.de}}
}
\maketitle

\begin{abstract}
Radar-based human sensing typically relies on end-to-end task-specific models trained on hand-chosen signal projections, which generalize poorly across unseen users, tasks, and hardware. We argue that the dominant signal processing choice, the micro-Doppler spectrogram, is a poor starting point for transfer because it keeps the Doppler axis, which carrier frequency rescales, and discards the angular axis, which carrier frequency leaves unchanged. We instead pretrain a convolutional $\beta$-VAE without labels on clutter-removed raw radar measurements from four platforms spanning 60 and 77\,GHz and 4 to 12 virtual antennas, aligned by a unified spectral interpolation grid. For every downstream benchmark the encoder stays frozen, and only a small supervised temporal classifier is trained on its embeddings. We compare against supervised models trained end to end on each benchmark and a pretrained radar foundation model across ten settings covering activity, gesture, sign-language, and gait recognition.
When the test users and the radar are seen during training, our encoder is best on all five benchmarks. For users never seen in training, it leads on three of four benchmarks, by $4.8$ to $23.2$ points. On subject-disjoint gait recognition, it is best for each of the four held-out subjects and $8.3$ points above the strongest baseline on average, even though its pretraining excluded every recording of this radar. Excluding the test radar from pretraining costs at most $5.3$ points on any benchmark. Code is available at \url{https://github.com/jonasmueler/UniversalRadarEncoder_WACV}.
\end{abstract}

\section{Introduction}
\label{sec:intro}
\begin{figure*}[!t]
  \centering
  \includegraphics[width=\textwidth]{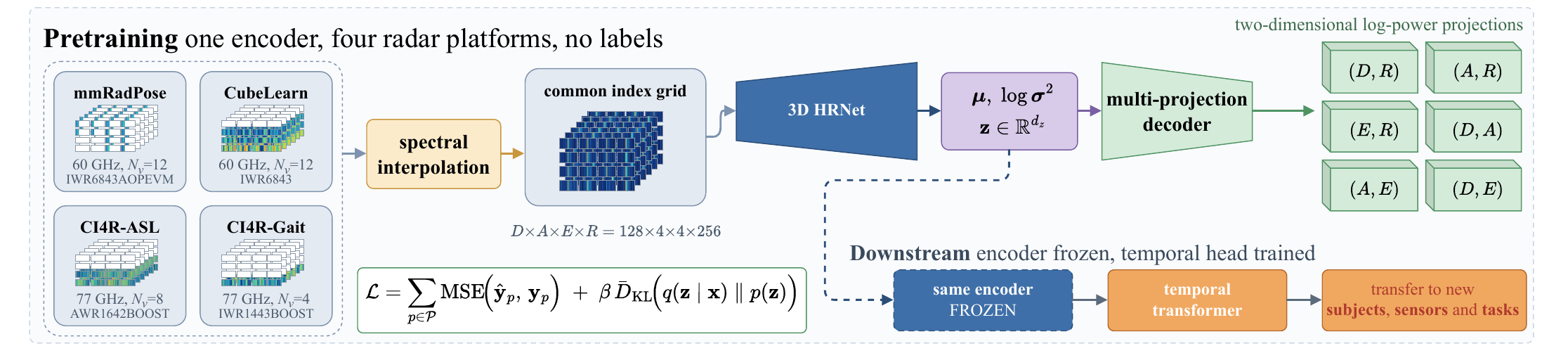}
  \caption{\textbf{One frozen encoder across four radar platforms.} A
  $\beta$-VAE is pretrained without labels on raw IQ from four sensors
  spanning 60 and 77\,GHz and 4 to 12 virtual antennas, aligned on a
  common index grid by spectral interpolation. The encoder is then frozen
  and reused unchanged, with only a small temporal head trained per
  benchmark, including under simultaneous shifts in subject, sensor,
  and task.}
  \label{fig:overview}
\end{figure*}

Radar-based contactless human sensing is gaining interest, from biomedical monitoring~\cite{krauss2024review} to human pose estimation~\cite{lee2023hupr, muller2026radproposer, mueller2024end, 10904474, mueller2026you, mueller2026learning} and activity recognition~\cite{ullmann2023radar, ullmann2023survey, ullmann2024classification, seyfiouglu2017deep, seyfiouglu2018deep}. Most models in this literature are trained and tested on one sensor. Radar hardware, however, differs substantially in antenna configuration, field of view, carrier frequency, and signal characteristics~\cite{10.1117/12.2559155, huang2025towards}, and representations do not carry over directly from one radar to another. Spectrogram classifiers for activity classification, for example, degrade severely when the test sensor differs from the training sensor~\cite{10.1117/12.2559155}, so transfer between sensors is a hard, open problem. We therefore pretrain one encoder across several radars, freeze it, and train only a small classifier on each benchmark's labels (\cref{fig:overview}), so that what transfers to a new sensor is the representation rather than the classifier itself.

Before learning, the signal is usually tailored to the task. A frequency-modulated continuous-wave (FMCW) MIMO radar delivers a complex-valued multi-dimensional tensor, the \emph{radar cube}, that combines range, velocity, and angular information from multiple synchronized transmit--receive antenna pairs~\cite{krauss2024review}. Most pipelines collapse this cube early into a lower-dimensional, task-specific representation, and these representations differ in density and in the information they preserve~\cite{krauss2024review}. Activity recognition sits at the dense, heavily projected end, where most models classify a single 2D time--frequency projection, above all the micro-Doppler spectrogram~\cite{ullmann2023survey, seyfiouglu2018deep, seyfiouglu2017deep, 10.1117/12.2559155}, which preserves the Doppler dynamics of human motion but collapses range and angle entirely. Pose-estimation and automotive pipelines usually sit at the sparse end, reducing the signal to point clouds via constant false alarm rate (CFAR) detection and angle-of-arrival (AoA) estimation~\cite{10904474, sun2020mimo}, which keeps an explicit representation suited to localization. Only a few systems operate close to the original tensor~\cite{zhao2023cubelearn, mueller2024end, muller2026radproposer, lee2023hupr}. For the input of such an encoder, new evidence favors data close to the raw signal. On a 3D occupancy-prediction task, omitting the Doppler FFT raises the test loss by 22.5\%, replacing dense antenna measurements with AoA estimates by 28.0\%, and applying CFAR thresholding by 31.5\%, each as large a degradation as training on more than ten times less data~\cite{huang2025towards}. Operating on the raw complex tensor~\cite{muller2026radproposer} improves pose estimation by 19\% over the CFAR point-cloud pipeline of Engel \etal~\cite{10904474}, and in activity recognition the spectrogram yields 86.49\% accuracy, the range--time map 95.25\%, and fusing multiple views of the cube 95.80\%~\cite{ullmann2023radar}. In general, a gradient in performance gains can be observed by including more of the raw data in the input to the encoder.

We also expect a physical advantage for transfer using the full tensor. As we derive in \cref{sec:radar-background}, the Doppler axis that the spectrogram keeps rescales with the carrier frequency, while the angular axis it discards does not on a half-wavelength array. Raw IQ, the complex-valued in-phase and quadrature (IQ) cube after a Fourier transform along all four axes, keeps both, since every measurement axis reaches the encoder uncollapsed. Whether one frozen encoder pretrained on raw IQ from several radars works on users and radars it has not seen remains open, and this is what our experiments test.

Two choices make joint pretraining across heterogeneous sensors possible. First, spectral interpolation and normalization align the radar tensors of all platforms on a common grid. Second, the encoder, a convolutional $\beta$-variational autoencoder ($\beta$-VAE) built on 3D High-Resolution Network (HRNet) layers~\cite{9052469, ho2024rt, muller2026radproposer}, takes the real and imaginary components as two channels. It is trained to reconstruct six 2D log-power projections rather than the input cube, because reconstructing the full 4D tensor is expensive and largely redundant, while fusing several projections improves classification accuracy in practice~\cite{ullmann2023radar}. Our contributions are:

\begin{itemize}
    \item A spectral-interpolation grid that aligns radar cubes of differing
    geometry, enabling joint pretraining across 60 and 77\,GHz single-chip
    platforms with 4 to 12 virtual antennas.
    \item A $\beta$-VAE that learns transferable representations from unlabeled
    raw IQ by reconstructing six log-power projections.
    \item A frozen-encoder evaluation over ten settings on these platforms, in
    which our representations lead the strongest supervised and pretrained
    baselines by up to $23.2$ points for unseen users and by up to $22.9$ points
    when the sensor is unseen as well.
\end{itemize}

\section{Related Work}
\label{sec:related}

\paragraph{Representations for human sensing with radar.}
Most FMCW human-sensing pipelines reduce the raw cube before any learning, to point clouds or to micro-Doppler spectrograms. For activity recognition, the standard recipe treats the micro-Doppler spectrogram as a 2D Doppler--time image and classifies it with a convolutional network~\cite{seyfiouglu2017deep, seyfiouglu2018deep, 10.1117/12.2559155}, which our Spectrogram CNN baseline reproduces. Working closer to the cube, Zhao \etal~\cite{zhao2023cubelearn} make the range--Doppler transform itself learnable and train an end-to-end classifier on the result. We retrain their strongest model as our supervised reference. Stadelmayer \etal~\cite{9464267} classify activities directly from raw ADC samples with parametric sinc and wavelet kernels, but their proprietary sensor and unavailable code and recordings prevent replication here.

\paragraph{General-purpose radar encoders.}
The closest comparison is GRT~\cite{huang2025towards}, a foundation model pretrained on one million 4D radar cubes from a single sensor against lidar-derived occupancy targets. It matches the single-chip geometry of our benchmarks and is available pretrained, so we evaluate it under our protocol. Other self-supervised radar work targets one problem each, for example gesture recognition~\cite{sheng2022facilitating}, human pose~\cite{wei2026maepose}, spatial uncertainty~\cite{Wang_2026_CVPR}, or micro-Doppler recognition~\cite{rahman2023self}, rather than an encoder reused across tasks and sensors.

\paragraph{Cross-sensor and cross-domain transfer.}
The classification performance of radar models shifts across environments~\cite{tanaka2025comprehensive}, subjects and positions~\cite{pinyoanuntapong2023gaitsada, lin2026active}, and previously unseen actions~\cite{xue2023towards}. Hardware heterogeneity is a particularly severe factor. Gurbuz \etal~\cite{10.1117/12.2559155} document an approximate $70\%$ performance collapse when the pretraining, fine-tuning, and evaluation platforms differ. Even the successful cross-frequency transfer of Shrestha \etal~\cite{shrestha2019cross}, which carries a 25\,GHz model to a 5.8\,GHz radar, relies on hand-crafted normalization of the spectrogram axis rather than a learned representation. Existing mitigation strategies fit models to each deployment through direct target measurement~\cite{pinyoanuntapong2023gaitsada, tanaka2025comprehensive}, unsupervised collection~\cite{ding2022source}, budget-constrained labeling~\cite{lin2026active}, or simulation~\cite{10.1117/12.2559155, 8572732, xue2023towards}. They also predominantly use micro-Doppler or time-Doppler spectrograms~\cite{10.1117/12.2559155, lin2026active, pinyoanuntapong2023gaitsada, 8572732, shrestha2019cross}. We instead learn directly from raw IQ and evaluate a fully frozen encoder on radar hardware excluded from pretraining, across antenna configurations and carrier frequencies.

\section{FMCW Radar Data Representation}
\label{sec:radar-background}

\paragraph{Goal and common grid.}
We seek one encoder that transfers to radar platforms excluded from its pretraining, without adapting its weights. The obstacle is that platforms differ in antenna count, sample count, sweep timing, and carrier frequency, so their raw tensors agree in neither shape nor spectral scaling. We therefore map every recording onto one canonical tensor, the common index grid, with $R{=}256$ fast-time samples, $D{=}128$ chirps, and a $4{\times}4$ antenna plane ($A{=}E{=}4$). Much like images from different cameras resized to one resolution, all sensors then share the same axes and origins after zero-filling and a Fourier transform, while each axis keeps a fixed scale per sensor, like a camera's zoom. On the receive axis this scale is practically identical across carriers, because each board spaces its receivers at about its own half wavelength, so on the row boards equal indices mean equal directions at 60 and 77\,GHz alike. On range and Doppler it remains a per-sensor constant, and the carrier adds a further stretch to Doppler from 60 to 77\,GHz. Since these constants never change with the scene, the encoder can learn to absorb them, as we argue below.

\begin{table*}[t]
    \centering
    \small
    \caption{Sensors and downstream benchmarks. HAR, HGR, AGR, and ASL denote human activity recognition, hand gesture recognition, arm gesture recognition, and American Sign Language, respectively. $N_v$ is the number of virtual antenna channels embedded in the common $4{\times}4$ grid.}
    \label{tab:datasets}
    \setlength{\tabcolsep}{5pt}
    \resizebox{\textwidth}{!}{%
    \begin{tabular}{lccccl}
        \toprule
        \textbf{Dataset} & \textbf{Platform} & \textbf{Carrier} & $\boldsymbol{N_v}$ & \textbf{Benchmark} & \textbf{Split and evaluation role} \\
        \midrule
        mmRadPose & IWR6843AOPEVM & 60\,GHz & 12 & 12 rehabilitation exercises & participant-disjoint, unseen-subject test \\
        CubeLearn & IWR6843ISK & 60\,GHz & 12 & HAR, HGR, AGR & in-set comparison, out-of-set unseen-user test \\
        CI4R-ASL & AWR1642BOOST & 77\,GHz & 8 & 18 daily-activity and ASL classes & subject-mixed, published-protocol comparison \\
        CI4R-Gait & IWR1443BOOST & 77\,GHz & 4 & 11 activities and gaits & subject-mixed comparison, leave-one-subject-out test \\
        \bottomrule
    \end{tabular}}
\end{table*}

\paragraph{The radar cube.}
An FMCW radar transmits short frequency sweeps, called chirps, and mixes each returning echo with the transmitted signal~\cite{richards2005fundamentals, 9658500}. Digitizing gives, per frame, a complex-valued tensor, the \emph{radar cube} $x[s, m, n_a, n_e]$, indexed by the sample $s$ within a chirp (fast time), the chirp $m$ (slow time), and two antenna axes $n_a, n_e$, on which a MIMO radar treats each transmit--receive pair as one \emph{virtual} element. For $K$ scatterers at range $r$, radial velocity $v$, and direction $(\theta, \phi)$, each axis carries one phase term,
\begin{equation}
  x[s, m, n_a, n_e] = \sum_{k=1}^{K} \alpha_k \,
  e^{\,\jmath 2\pi \left( f^R_k s \,+\, f^D_k m \,+\, \psi^A_k(n_a) \,+\, \psi^E_k(n_e) \right)},
  \label{eq:cube}
\end{equation}
where $\alpha_k$ is the complex reflectivity of scatterer $k$~\cite{9658500}. Along fast time, an echo from range $r$ returns after a delay $2r/c$, with light speed $c$, so mixing with a sweep of slope $S$ leaves $f^R = 2Sr/(c f_s)$ cycles per sample at sampling rate $f_s$. Along slow time, a scatterer moves by $vT_c$ during the per-transmitter chirp interval $T_c$, advancing its phase by $f^D = 2T_c v/\lambda$ cycles per chirp at wavelength $\lambda$. Across the array, the phase is the extra path to a virtual antenna in wavelengths. It splits into one term per antenna axis, $\psi^A$ and $\psi^E$. On mmRadPose (\cref{tab:datasets}) these follow azimuth and elevation, on the other boards $\psi^A$ follows the transmitters and $\psi^E$ the receivers. A Fourier transform along an axis turns a linear phase into a peak at an index proportional to its frequency.

\paragraph{Carrier-frequency invariant structure.}
Pretraining across radars depends on what an index means on each sensor, and \cref{eq:cube} answers this axis by axis. Range is carrier-invariant, since $f^R$ contains no wavelength. The receive axis is carrier-invariant as well, because each board spaces its receivers at about its own $\lambda/2$, in a row on three boards and in 2D on the IWR6843AOP of mmRadPose. Element $n_e$ of a row then has phase $\tfrac{1}{2} n_e \sin\theta$, the wavelength cancels, and on the row boards the same $E$ index means the same azimuth at 60 and 77\,GHz, while on mmRadPose both array axes are carrier-free. Only Doppler keeps the carrier, through $\lambda$ in $f^D$. The encoder must learn to compensate for the remaining sensor-specific differences. On range and Doppler, sensors differ by a stretch that maps zero to zero, set by sweep slope and sampling rate for range and by chirp timing and wavelength for Doppler. On the antenna axes, each sensor contributes a fixed pattern. Time-division (mmRadPose, CubeLearn) and phase-coded (CI4R-ASL) boards also leave an uncompensated velocity-dependent phase or leakage between transmitter streams. Stretches and patterns are hardware constants, and this residual is a fixed per-sensor function of velocity, which makes joint pretraining plausible for arrays with a half-wavelength receive axis, while other receive geometries would need resampling or calibration. Our leave-one-sensor-out experiments test whether a frozen encoder absorbs these differences without adaptation.

\paragraph{Preprocessing onto the common grid.}

\begin{figure}[b]
  \centering
  \includegraphics[width=\columnwidth]{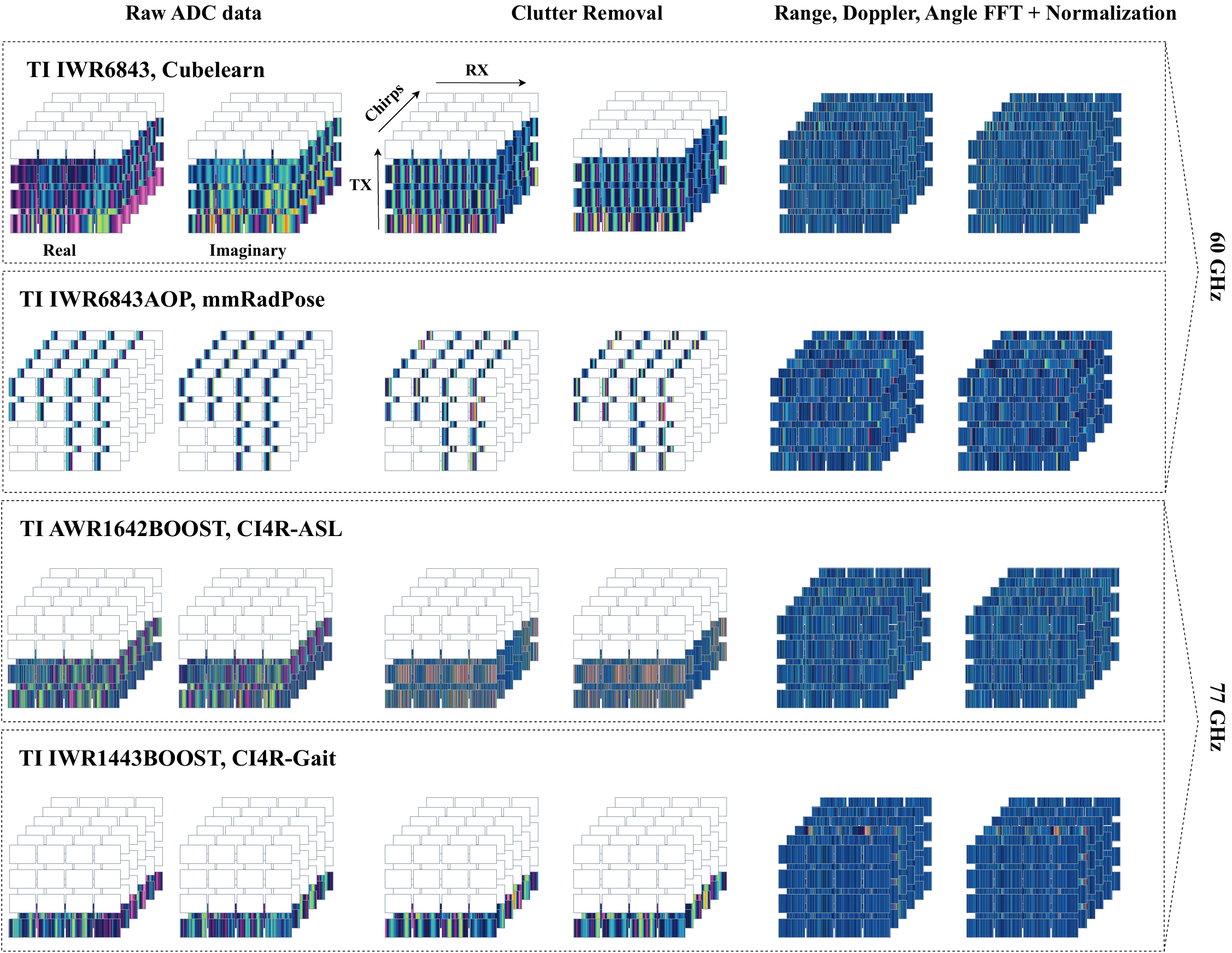}
  \caption{Preprocessing on real recordings from all four sensors. Each block is one frame on the canonical $D{\times}A{\times}E{\times}R$ grid, with six chirp slices, the $4{\times}4$ antenna plane as cells, and range across each cell. Real and imaginary parts are shown separately. Outlined cells are zero-filled. The sensors populate 12, 12, 8, and 4 of the 16 antenna cells, respectively, and mmRadPose fills 64 of 256 range samples. The transform chain aligns range and Doppler while retaining each sensor's fixed antenna pattern.}
  \label{fig:preprocessing}
\end{figure}

Four deterministic steps map every sensor to the common grid (\cref{fig:preprocessing}). \emph{(i)} Static clutter is removed by subtracting from every antenna--sample series its mean over the chirps of the frame, which suppresses the zero-Doppler component and keeps the moving returns. \emph{(ii)} The cube is zero-filled into the canonical grid, which is the spectral interpolation of our pipeline and discards no sample. The twelve virtual antennas of the prevailing three-transmitter, four-receiver configuration are written into the $4{\times}4$ plane by a fixed element-to-cell map, sensors with fewer chains populate fewer cells, and, since $R$ and $D$ match or exceed every platform's counts, no recording is truncated. \emph{(iii)} A Fourier transform is applied along all four axes. \emph{(iv)} Each frame is normalized to unit RMS magnitude to remove sensor gain and enters the model as two real channels holding its real and imaginary parts.

\section{Method}
\label{sec:method}

\begin{table*}[t]
    \centering
    \small
    \caption{CubeLearn under the official protocol of~\cite{zhao2023cubelearn},
    where in-set = seen users 0--5 and out = unseen users 6--7. Mean$_{\pm\text{std}}$
    over 3 seeds. \textbf{Bold} = best per column and panel.}
    \label{tab:cube_comparison}
    \setlength{\tabcolsep}{4pt}
    \begin{tabular}{lcccccc}
        \toprule
        & \multicolumn{2}{c}{\textbf{Cube:HAR}} & \multicolumn{2}{c}{\textbf{Cube:HGR}} & \multicolumn{2}{c}{\textbf{Cube:AGR}} \\
        \cmidrule(lr){2-3} \cmidrule(lr){4-5} \cmidrule(lr){6-7}
        \textbf{Model} & in-set & out & in-set & out & in-set & out \\
        \midrule
        \multicolumn{7}{l}{\textit{(a) Accuracy}} \\
        \midrule
        RDAT (DFT)~\cite{zhao2023cubelearn} & $0.992_{\pm0.003}$ & $0.721_{\pm0.080}$ & $0.973_{\pm0.001}$ & $0.491_{\pm0.030}$ & $0.986_{\pm0.011}$ & $0.784_{\pm0.013}$ \\
        RDAT (CubeLearn)~\cite{zhao2023cubelearn} & $0.992_{\pm0.006}$ & $0.765_{\pm0.103}$ & $0.969_{\pm0.005}$ & $0.491_{\pm0.043}$ & $0.991_{\pm0.008}$ & $0.847_{\pm0.016}$ \\
        RadMamba~\cite{11316534} & $0.930_{\pm0.050}$ & $0.865_{\pm0.097}$ & $0.744_{\pm0.042}$ & $0.629_{\pm0.053}$ & $0.929_{\pm0.009}$ & $0.810_{\pm0.011}$ \\
        GRT~\cite{huang2025towards} & $0.994_{\pm0.006}$ & $0.468_{\pm0.028}$ & $0.969_{\pm0.005}$ & $0.696_{\pm0.009}$ & $0.990_{\pm0.003}$ & $\mathbf{0.964}_{\pm0.016}$ \\
        Spectrogram CNN & $0.981_{\pm0.006}$ & $0.882_{\pm0.004}$ & $0.719_{\pm0.026}$ & $0.602_{\pm0.024}$ & $0.955_{\pm0.001}$ & $0.863_{\pm0.013}$ \\
        \midrule[\heavyrulewidth]
        \multicolumn{7}{l}{\textbf{Ours}} \\
        Raw-IQ VAE (all sensors) & $\mathbf{0.995}_{\pm0.006}$ & $\mathbf{0.930}_{\pm0.034}$ & $\mathbf{0.980}_{\pm0.005}$ & $\mathbf{0.825}_{\pm0.005}$ & $\mathbf{0.996}_{\pm0.006}$ & $0.905_{\pm0.008}$ \\
        Raw-IQ VAE (sensor held out) & $0.992_{\pm0.012}$ & $0.877_{\pm0.051}$ & $0.957_{\pm0.006}$ & $0.784_{\pm0.015}$ & $\mathbf{0.996}_{\pm0.003}$ & $0.956_{\pm0.007}$ \\
        \midrule
        \multicolumn{7}{l}{\textit{(b) Macro-F1}} \\
        \midrule
        RDAT (DFT)~\cite{zhao2023cubelearn} & $0.992_{\pm0.003}$ & $0.710_{\pm0.083}$ & $0.973_{\pm0.001}$ & $0.475_{\pm0.023}$ & $0.986_{\pm0.011}$ & $0.780_{\pm0.018}$ \\
        RDAT (CubeLearn)~\cite{zhao2023cubelearn} & $0.992_{\pm0.006}$ & $0.754_{\pm0.103}$ & $0.969_{\pm0.005}$ & $0.477_{\pm0.034}$ & $0.991_{\pm0.008}$ & $0.845_{\pm0.015}$ \\
        RadMamba~\cite{11316534} & $0.929_{\pm0.050}$ & $0.858_{\pm0.104}$ & $0.742_{\pm0.042}$ & $0.628_{\pm0.053}$ & $0.929_{\pm0.009}$ & $0.808_{\pm0.011}$ \\
        GRT~\cite{huang2025towards} & $0.994_{\pm0.006}$ & $0.459_{\pm0.039}$ & $0.970_{\pm0.005}$ & $0.676_{\pm0.003}$ & $0.989_{\pm0.003}$ & $\mathbf{0.964}_{\pm0.016}$ \\
        Spectrogram CNN & $0.981_{\pm0.006}$ & $0.879_{\pm0.005}$ & $0.715_{\pm0.027}$ & $0.594_{\pm0.020}$ & $0.955_{\pm0.001}$ & $0.859_{\pm0.016}$ \\
        \midrule[\heavyrulewidth]
        \multicolumn{7}{l}{\textbf{Ours}} \\
        Raw-IQ VAE (all sensors) & $\mathbf{0.995}_{\pm0.006}$ & $\mathbf{0.928}_{\pm0.035}$ & $\mathbf{0.980}_{\pm0.005}$ & $\mathbf{0.812}_{\pm0.003}$ & $\mathbf{0.996}_{\pm0.006}$ & $0.904_{\pm0.008}$ \\
        Raw-IQ VAE (sensor held out) & $0.991_{\pm0.013}$ & $0.878_{\pm0.051}$ & $0.957_{\pm0.006}$ & $0.777_{\pm0.018}$ & $\mathbf{0.996}_{\pm0.003}$ & $0.956_{\pm0.006}$ \\
        \bottomrule
    \end{tabular}

\end{table*}

Our method has two phases. During unlabeled pretraining, an encoder maps each aligned IQ frame to a latent code, and a set of decoders reconstructs six power projections. The decoders are then discarded. For every downstream benchmark, the encoder remains frozen, each recording becomes a sequence of per-frame posterior means, and a small temporal classifier is trained on the benchmark's labels.

\subsection{Model architecture}
\label{sec:architecture}

The pretraining model is a $\beta$-variational autoencoder ($\beta$-VAE)~\cite{kingma2013auto, higgins2017beta} that maps one radar frame to a compact latent vector and reconstructs a set of two-dimensional power projections.

\paragraph{Encoder.}
The IQ input from \cref{sec:radar-background} has shape $(D, A, E, R, C)$, with $D{=}128$ chirps, $A{=}E{=}4$ antenna dimensions, $R$ range bins, and $C{=}2$ real and imaginary channels. To use a 3D backbone, we reshape the two IQ channels into the Doppler extent, giving a volume of size $(D{\times}C, A, E, R)$, which preserves every input value. A 3D HRNet~\cite{9052469, ho2024rt, muller2026radproposer} extracts multi-resolution features, which a two-layer transformer aggregates through a learned \texttt{[CLS]} token. Two linear heads produce the mean $\boldsymbol{\mu}$ and log-variance $\log \boldsymbol{\sigma}^2$ of a diagonal Gaussian with $d_z{=}256$. During training, we sample $\mathbf{z} = \boldsymbol{\mu} + \boldsymbol{\sigma} \odot \boldsymbol{\epsilon}$ with $\boldsymbol{\epsilon} \sim \mathcal{N}(\mathbf{0}, \mathbf{I})$.

\paragraph{Multi-projection decoder.}
Rather than reconstructing the full four-dimensional cube, the decoders predict six two-dimensional log-power projections. For each retained pair of axes, we sum IQ power over the other two, giving range--Doppler~$(D, R)$, range--azimuth~$(A, R)$, range--elevation~$(E, R)$, Doppler--azimuth~$(D, A)$, azimuth--elevation~$(A, E)$, and Doppler--elevation~$(D, E)$. Each map has a lightweight decoder sized to its output resolution. Thus, the encoder receives the full complex tensor, while projection-space targets provide efficient supervision that preserves pairwise structure across range, velocity, and angle without full 4D reconstruction.

\paragraph{Downstream classifier.}
\label{sec:classifier}
Each recording becomes a variable-length sequence of per-frame embeddings $(\mathbf{e}_1,\dots,\mathbf{e}_T)$, where $\mathbf{e}_t = \boldsymbol{\mu}_t \in \mathbb{R}^{d_z}$ is the latent mean of frame $t$. A temporal transformer with a learned \texttt{[CLS]} token, positional embeddings, and a padding mask maps that sequence to class logits. Within each benchmark, every encoder is evaluated with the identical head and training protocol, apart from its input width and the number of classes. The head has one layer on mmRadPose and CubeLearn and four layers on CI4R, whose recordings are an order of magnitude longer, with dropout~0.3 throughout. The head reads the deterministic mean at both training and test time and is trained for 200~epochs with cross-entropy and label smoothing of~0.1. The validation-selected checkpoint is evaluated once on the held-out test split.

\subsection{Training}
\label{sec:training}

We train the $\beta$-VAE~\cite{kingma2013auto, higgins2017beta} by minimizing the negative $\beta$-weighted evidence lower bound of the projections, estimated with one reparameterized sample per frame,
\begin{equation}
  \mathcal{L} = \sum_{p \in \mathcal{P}} \mathrm{MSE}\!\big(\hat{\mathbf{y}}_p,\, \mathbf{y}_p\big)
  \;+\; \beta\, \bar{D}_{\mathrm{KL}}\!\big(q(\mathbf{z} \mid \mathbf{x}) \;\|\; p(\mathbf{z})\big),
  \label{eq:loss}
\end{equation}
where $\mathcal{P}$ is the set of six projections, $\hat{\mathbf{y}}_p$ and $\mathbf{y}_p$ are the predicted and target log-power maps, and $\bar{D}_{\mathrm{KL}} = D_{\mathrm{KL}}/d_z$ is normalized per latent dimension. Element-wise MSE gives every projection equal weight regardless of its size, while the KL term regularizes the latent posterior toward $p(\mathbf{z}) = \mathcal{N}(\mathbf{0}, \mathbf{I})$. We set $\beta{=}0.1$ with a five-epoch linear warmup and train with AdamW, cosine annealing, gradient clipping at 1.0, batch size 16, and 50 fixed-length epochs of frames drawn with replacement. Each dataset receives an equal share, $1/4$ in the all-sensor pool and $1/3$ in each leave-one-sensor-out pool, with CubeLearn's share split evenly across HAR, HGR, and AGR, so every pool receives the same number of optimization steps.

\subsection{Datasets}
\label{sec:datasets}

\begin{table}[b]
    \centering
    \small
    \caption{mmRadPose, 12-class participant-disjoint HAR, chance ${=}0.083$.
    Mean$_{\pm\text{std}}$ over 3 seeds. \textbf{Bold} = best per column.}
    \label{tab:mmrad_comparison}
    \setlength{\tabcolsep}{5pt}
    \resizebox{\columnwidth}{!}{%
    \begin{tabular}{lcc}
        \toprule
        \textbf{Model} & \textbf{Accuracy} & \textbf{Macro-F1} \\
        \midrule
        RDAT (DFT)~\cite{zhao2023cubelearn} & $0.088_{\pm0.005}$ & $0.013_{\pm0.001}$ \\
        RDAT (CubeLearn)~\cite{zhao2023cubelearn} & $0.091_{\pm0.005}$ & $0.014_{\pm0.001}$ \\
        RadMamba~\cite{11316534} & $0.484_{\pm0.039}$ & $0.473_{\pm0.045}$ \\
        GRT~\cite{huang2025towards} & $0.648_{\pm0.088}$ & $0.629_{\pm0.101}$ \\
        Spectrogram CNN & $0.525_{\pm0.005}$ & $0.515_{\pm0.009}$ \\
        \midrule[\heavyrulewidth]
        \multicolumn{3}{l}{\textbf{Ours}} \\
        Raw-IQ VAE (all sensors) & $\mathbf{0.880}_{\pm0.077}$ & $0.870_{\pm0.089}$ \\
        Raw-IQ VAE (sensor held out) & $0.877_{\pm0.016}$ & $\mathbf{0.877}_{\pm0.016}$ \\
        \bottomrule
    \end{tabular}}

\end{table}

Four public FMCW radar datasets serve as unlabeled pretraining sources and labeled downstream benchmarks (\cref{tab:datasets}). Wherever a release provides a participant-disjoint protocol, we use it, because it directly tests transfer to a new person. CI4R-ASL and CI4R-Gait publish subject-mixed protocols, which we keep for protocol-matched comparison with prior work, and we add leave-one-subject-out evaluation on CI4R-Gait to isolate subject transfer. Every pretraining pool contains only non-test data, and each downstream test set stays unseen during pretraining and classifier training. The held-out-sensor claims use leave-one-sensor-out encoders whose pool excludes the evaluated platform entirely (\cref{sec:loseno}).

\begin{table*}[t]
    \centering
    \small
    \caption{The two 77\,GHz CI4R benchmarks under their releases' subject-mixed protocols, namely CI4R-ASL~\cite{9660776} (18 activity and sign classes, chance ${=}0.056$) and CI4R-Gait~\cite{10.1117/12.2559155} (11 activities and gaits, chance ${=}0.091$). Mean$_{\pm\text{std}}$ over 3 seeds. Every row is scored on one identical partition per release, so all rows are directly comparable. \textbf{Bold} = best per column.}
    \label{tab:ci4r_comparison}
    \setlength{\tabcolsep}{4pt}
    \begin{tabular*}{\textwidth}{@{\extracolsep{\fill}}lcccc}
        \toprule
        & \multicolumn{2}{c}{\textbf{CI4R-ASL}} & \multicolumn{2}{c}{\textbf{CI4R-Gait}} \\
        \cmidrule(lr){2-3} \cmidrule(lr){4-5}
        \textbf{Model} & Acc. & Macro-F1 & Acc. & Macro-F1 \\
        \midrule
        GRT~\cite{huang2025towards} & $0.938_{\pm0.010}$ & $0.899_{\pm0.015}$ & $0.915_{\pm0.023}$ & $0.927_{\pm0.020}$ \\
        RDAT (DFT)~\cite{zhao2023cubelearn} & $0.737_{\pm0.036}$ & $0.535_{\pm0.064}$ & $0.675_{\pm0.051}$ & $0.696_{\pm0.060}$ \\
        RDAT (CubeLearn)~\cite{zhao2023cubelearn} & $0.679_{\pm0.019}$ & $0.444_{\pm0.031}$ & $0.633_{\pm0.025}$ & $0.632_{\pm0.041}$ \\
        RadMamba~\cite{11316534} & $0.934_{\pm0.006}$ & $0.890_{\pm0.010}$ & $0.810_{\pm0.023}$ & $0.815_{\pm0.032}$ \\
        Spectrogram CNN & $0.951_{\pm0.003}$ & $0.918_{\pm0.004}$ & $0.723_{\pm0.008}$ & $0.716_{\pm0.007}$ \\
        \midrule[\heavyrulewidth]
        \multicolumn{5}{l}{\textbf{Ours}} \\
        Raw-IQ VAE (all sensors) & $\mathbf{0.971}_{\pm0.007}$ & $\mathbf{0.953}_{\pm0.010}$ & $\mathbf{0.951}_{\pm0.012}$ & $\mathbf{0.953}_{\pm0.011}$ \\
        Raw-IQ VAE (sensor held out) & $0.967_{\pm0.002}$ & $0.946_{\pm0.003}$ & $0.900_{\pm0.020}$ & $0.906_{\pm0.015}$ \\
        \bottomrule
    \end{tabular*}

\end{table*}

\paragraph{mmRadPose.}
mmRadPose~\cite{10904474, muller2026radproposer} holds raw ADC recordings of 12~participants performing 12~rehabilitation exercises. We use its participant-disjoint protocol~\cite{muller2026radproposer}, the release's direct test of generalization to a new person, with p4--p11 for training, p3 for validation, and p1, p2, and p12 for testing. Both the VAE and the classifier train on p4--p11 and select their checkpoints on p3, and the classifier is evaluated once on the three test participants, which neither stage has seen.

\paragraph{CubeLearn.}
CubeLearn~\cite{zhao2023cubelearn} holds raw ADC data of 8~users in HAR, hand-gesture (HGR), and arm-gesture (AGR) subsets. Its in-set split, the standard in-distribution comparison, divides the recording indices of users 0--5 into 15/5/10 train, validation, and test samples. The out-of-set split withholds users 6--7 entirely and tests on all 30 of their samples per user and class, giving 360 recordings on HAR and 720 on each gesture task, and is our second participant-disjoint test. The VAE uses only in-set train and validation frames, without labels, so no component sees an out-of-set user. HGR and AGR contribute their training splits to the all-sensor pool and, under leave-one-sensor-out, are unseen as both task and sensor.

\paragraph{CI4R-ASL.}
CI4R-ASL is the ASL Sequential Dataset of~\cite{9660776}. The benchmark classifies ground-truth segments, 1496~labeled repetitions in 324~recordings, split into 1047 training, 150 validation, and 299 test repetitions. We keep the authors' shuffled subject-mixed protocol, the only published setting for a protocol-matched comparison. Since every participant appears in training and test, the comparison is protocol-matched rather than evidence of cross-user transfer. Only the 1047 training repetitions enter any pretraining pool, and the VAE carves its own validation slice from them, so benchmark validation and test repetitions enter neither pretraining nor classifier training.

\paragraph{CI4R-Gait.}
CI4R-Gait is the activity-and-gait release of~\cite{10.1117/12.2559155}, in which six participants perform 11~daily activities and ambulatory gaits in 652~recordings under five released subject codes, which we treat as subjects. We use only its 77\,GHz channel and keep the shuffled subject-mixed protocol, with 457 training, 65 validation, and 130 test recordings, for a protocol-matched comparison with its published spectrogram classifier. Only the 457 training recordings enter any pretraining pool, again with the VAE's validation slice carved from them. Because subject mixing cannot test person-independent recognition, we add leave-one-subject-out folds (\cref{sec:loso}). Four usable subject codes contribute 116, 91, 221, and 214~recordings, and a fifth with ten single-class recordings always stays in training, which fixes the evaluation at four folds.

\section{Experimental Setup}
\label{sec:setup}
We evaluate two kinds of encoder. The all-sensor encoder is pretrained without labels on all four platforms and every task. Each leave-one-sensor-out encoder drops one platform from that pool and is evaluated on the dropped platform's benchmark, so its test sensor contributed no frame to its pretraining. Pretraining draws only on the training and validation frames of each benchmark, and a held-out subject is withheld from it as well as from the classifier. Every number is mean$_{\pm\text{std}}$ over three seeds, reported as accuracy and macro-F1.

\begin{table*}[t]
    \centering
    \small
    \caption{Leave-one-subject-out CI4R-Gait, where each fold holds one subject code out
    of training, validation, and model selection. Per-fold cells are accuracy
    with held-out recordings beneath, and the last two columns average over folds.
    All cells are mean$_{\pm\text{std}}$ over 3
    seeds. \textbf{Bold} = best per column.}
    \label{tab:loso}
    \setlength{\tabcolsep}{4pt}
    \begin{tabular*}{\textwidth}{@{\extracolsep{\fill}}lcccccc}
        \toprule
        & \multicolumn{4}{c}{\textbf{Per-fold accuracy}} & \multicolumn{2}{c}{\textbf{Mean over folds}} \\
        \cmidrule(lr){2-5} \cmidrule(lr){6-7}
        \multirow{2}{*}{\textbf{Model}} & S1 & S2 & S3 & S4 & \textbf{Accuracy} & \textbf{Macro-F1} \\
        & \footnotesize 116 & \footnotesize 91 & \footnotesize 221 & \footnotesize 214 & \footnotesize 642 & \\
        \midrule
        GRT~\cite{huang2025towards} & $0.491_{\pm0.026}$ & $0.484_{\pm0.094}$ & $0.549_{\pm0.019}$ & $0.606_{\pm0.010}$ & $0.532_{\pm0.027}$ & $0.488_{\pm0.020}$ \\
        RDAT (DFT)~\cite{zhao2023cubelearn} & $0.474_{\pm0.106}$ & $0.509_{\pm0.022}$ & $0.439_{\pm0.012}$ & $0.540_{\pm0.026}$ & $0.490_{\pm0.030}$ & $0.447_{\pm0.026}$ \\
        RDAT (CubeLearn)~\cite{zhao2023cubelearn} & $0.368_{\pm0.057}$ & $0.381_{\pm0.017}$ & $0.395_{\pm0.017}$ & $0.422_{\pm0.085}$ & $0.392_{\pm0.012}$ & $0.358_{\pm0.024}$ \\
        RadMamba~\cite{11316534} & $0.575_{\pm0.088}$ & $0.593_{\pm0.038}$ & $0.603_{\pm0.025}$ & $0.670_{\pm0.035}$ & $0.610_{\pm0.028}$ & $0.587_{\pm0.026}$ \\
        Spectrogram CNN & $0.489_{\pm0.018}$ & $0.611_{\pm0.013}$ & $0.587_{\pm0.011}$ & $0.646_{\pm0.019}$ & $0.583_{\pm0.009}$ & $0.557_{\pm0.009}$ \\
        \midrule[\heavyrulewidth]
        \multicolumn{7}{l}{\textbf{Ours}} \\
        Raw-IQ VAE (sensor held out) & $\mathbf{0.635}_{\pm0.018}$ & $\mathbf{0.696}_{\pm0.034}$ & $\mathbf{0.679}_{\pm0.039}$ & $\mathbf{0.762}_{\pm0.005}$ & $\mathbf{0.693}_{\pm0.012}$ & $\mathbf{0.666}_{\pm0.010}$ \\
        \bottomrule
    \end{tabular*}

\end{table*}

\subsection{Baselines}
\label{sec:baselines}

We compare against three published methods, each run from its authors' released code under our splits and seeds. One rule governs all inputs. Every method runs in the configuration its authors published, receives clutter removal only if its authors used it, and is never given range bins cherrypicked for a particular dataset. RDAT\_3DCNNLSTM~\cite{zhao2023cubelearn}, the supervised model of CubeLearn, is evaluated with its conventional DFT and its learnable CubeLearn front end, denoted RDAT (DFT) and RDAT (CubeLearn), and keeps its published crop of 64 chirps $\times$ 128 range samples, because its LSTM input size is tied to that crop. RadMamba~\cite{11316534}, a supervised micro-Doppler model, runs the authors' spectrogram pipeline identically on every benchmark, without clutter removal and over the full range axis instead of their sensor-specific range window. GRT~\cite{huang2025towards}, a pretrained radar encoder foundation model, is frozen at its released checkpoint, receives every sensor on the Doppler--range grid and virtual-array layout it was pretrained on, and is classified with our temporal head. As the standard micro-Doppler pipeline, a Spectrogram CNN classifies the Doppler--time spectrogram of one receive channel over the whole recording, computed with clutter removal, with a 2D CNN trained from scratch on each benchmark. The published systems of the two CI4R releases~\cite{9660776, 10.1117/12.2559155} used train/test partitions whose seeds are unpublished, so we cite rather than tabulate them.

\section{Results}
\label{sec:results}

\subsection{No domain shift}
\label{sec:indist}

On the three CubeLearn in-set tasks, our encoder is on par with or above the best supervised model, with the largest margin on 12-class hand gestures ($0.980$ vs.\ $0.973$). Activity and arm-gesture recognition saturate at or above $0.99$ for the best model of each paradigm (\cref{tab:cube_comparison}). On the subject-mixed CI4R protocols, our encoder reaches $0.971$ on ASL and $0.951$ on Gait, $2.0$~points above the Spectrogram CNN and $3.6$~points above GRT, the strongest baselines on the two partitions (\cref{tab:ci4r_comparison}). The systems published with the two releases report $0.935$ on the hearing participants of ASL~\cite{9660776} and $0.915$ on Gait~\cite{10.1117/12.2559155}, on partitions drawn with unpublished seeds. On the hearing subset of our partition, our encoder reaches $0.982$. Without a domain shift, the strongest models are therefore within a few points of each other.

\subsection{Unseen subjects}
\label{sec:oneheldout}

CubeLearn's out-of-set protocol evaluates on users 6 and 7, who are absent from training. On these users, our all-sensor encoder is the best model for activities ($0.930$, $4.8$~points above the Spectrogram CNN) and hand gestures ($0.825$, $12.9$~points above GRT), and GRT is the best model for arm gestures, with $0.964$ against our $0.905$ (\cref{tab:cube_comparison}). From in-set to out-of-set, our encoder loses $6.5$ to $15.5$~points, whereas the two RDAT variants lose $14$ to $48$, consistent with the identity dependence the CubeLearn authors observed for their supervised models~\cite{zhao2023cubelearn}. On the participant-disjoint mmRadPose benchmark, our encoder is the best model with $0.880$, ahead of GRT ($0.648$) and the Spectrogram CNN ($0.525$), while both RDAT variants remain at chance level (\cref{tab:mmrad_comparison}). With new subjects, the frozen encoder therefore loses less accuracy than RDAT and is the best model on three of the four benchmarks, although only its temporal classifier is trained on the target benchmark, whereas RDAT and RadMamba train their entire networks on it.

\subsection{Unseen sensors}
\label{sec:loseno}
Each leave-one-sensor-out encoder is pretrained without any recording of the evaluated platform. Every platform was recorded in its own environment with its own label set, so holding out a sensor also withholds its motions from pretraining, and sensor and task change together throughout this section. As in all experiments, the temporal head is trained on the labeled training split of the target benchmark. Relative to the all-sensor encoder, holding the sensor out costs at most $5.3$~points (Cube:HAR out-of-set).

\paragraph{Subject-mixed protocols.}
On CubeLearn in-set, CI4R-ASL, and CI4R-Gait, the test subjects appear in training. Here the held-out encoder outperforms all baselines on CI4R-ASL ($0.967$ against $0.951$ for the Spectrogram CNN) and on CubeLearn arm gestures ($0.996$). GRT is ahead on CI4R-Gait ($0.915$ vs.\ $0.900$) and on CubeLearn activities ($0.994$ vs.\ $0.992$), and RDAT (DFT) on hand gestures ($0.973$ vs.\ $0.957$) (\cref{tab:ci4r_comparison,tab:cube_comparison}). With its sensor in pretraining, our encoder is the best model on all five subject-mixed benchmarks. Without it, the encoder still leads on two and trails the best model by at most $1.6$~points on the other three, although it has seen no recording of the target sensor before its classifier is trained, whereas RDAT and RadMamba learn their entire networks on that sensor's data.

\paragraph{Subject-disjoint protocols.}
\label{sec:loso}
On mmRadPose, CubeLearn out-of-set, and the CI4R-Gait folds below, no test subject appears in training. On mmRadPose, the held-out encoder is the best model with $0.877$, $22.9$~points above GRT (\cref{tab:mmrad_comparison}). On CubeLearn's unseen users, it outperforms all baselines on hand gestures ($0.784$, $8.8$~points above GRT), while the Spectrogram CNN is best on activities with $0.882$ against $0.877$ and GRT on arm gestures with $0.964$ against $0.956$ (\cref{tab:cube_comparison}). Of the two CI4R releases, only CI4R-Gait permits a subject-disjoint evaluation (\cref{sec:datasets}). We run it as four leave-one-subject-out folds with the encoder pretrained without CI4R-Gait, each withholding one subject from training, validation, and model selection (\cref{tab:loso}). The second fold lacks one activity, so its macro-F1 covers ten classes. Our encoder is the best model on every fold and reaches $0.693$ on average, $8.3$~points above RadMamba. Relative to the subject-mixed protocol, our accuracy drops by $20.7$~points and that of GRT by $38.3$, which places GRT below RadMamba. When the sensor and the test subjects are both new, the held-out encoder is therefore ahead of all baselines on three of the five benchmarks, with the Spectrogram CNN ahead on activities and GRT on arm gestures, even though RDAT and RadMamba train their entire networks on the target sensor's data and the encoder has seen no recording of that sensor before its classifier is trained.

\begin{table}[t]
    \centering
    \footnotesize
    \setlength{\tabcolsep}{4pt}
    \caption{Parameters (encoder and head, or the end-to-end model), forward GFLOPs (model forward only) per recording on CI4R-ASL, the largest benchmark, with recordings of intermediate length (69 frames on average), and input values per parameter (Val./Par.).}
    \label{tab:compute}
    \begin{tabular*}{\columnwidth}{@{\extracolsep{\fill}}lrrr}
        \toprule
        \textbf{Model} & Params (M) & GFLOPs & Val./Par. \\
        \midrule
        RDAT & $25.252$ & $28.39$ & $0.36$ \\
        RadMamba & $0.048$ & $0.0035$ & $1.04$ \\
        Spectrogram CNN & $0.393$ & $0.25$ & $0.09$ \\
        GRT & $25.886$ & $5{,}967$ & $1.40$ \\
        Raw-IQ VAE & $41.832$ & $29{,}534$ & $1.73$ \\
        \bottomrule
    \end{tabular*}
\end{table}

\paragraph{Computational cost.}
\label{sec:cost}
Our encoder is the most expensive model in parameters and forward cost (\cref{tab:compute}), because it reads the full raw cube of every frame. RadMamba and the Spectrogram CNN compress each recording into one image and are far cheaper, but trail our encoder on most domain-shift benchmarks. Among the models that read the radar cube, model size follows the input. To relate the two, we compute the ratio of the input values a model processes over one recording, after its own preprocessing, to its parameter count. On CI4R-ASL, our encoder processes $1.73$ input values per parameter, GRT $1.40$, and RDAT $0.36$, so our encoder needs about five times fewer parameters per input value than RDAT and fewer than GRT. Adapting it to a new benchmark trains only the temporal head. Encoding a recording on an NVIDIA A40 with PyTorch's default precision takes $26$\,ms per frame for our encoder and $7.7$\,ms for GRT. As GPU throughput and memory continue to grow, we consider this cost worth paying for the full resolution of the raw cube and the higher accuracy under domain shift, especially since pretraining is paid for only once.

\section{Limitations}
\label{sec:limitations}
Our experiments attribute the gains to the combination of our design choices rather than to any individual one. The common grid and its $4{\times}4$ angular resolution, the encoder architecture and objective, multi-sensor pretraining, and preprocessing such as clutter removal all differ between our encoder and the baselines. The benchmarks are single-person and short-range.

\section{Conclusion}
\label{sec:conclusion}
We pretrain a $\beta$-VAE on unlabeled raw radar IQ data from four platforms and keep it frozen afterwards. A small temporal head trained on its embeddings then suffices for activity, gesture, sign-language, and gait recognition. In distribution it matches supervised end-to-end training, and on unseen subjects it is substantially more accurate. When the sensor is also held out from pretraining, it still outperforms all baselines on hand gestures, on mmRadPose, and on subject-disjoint gait recognition, and remains within one point of the best baseline on arm gestures and activities. These results show that a simple VAE can be pretrained jointly on radar IQ data from several sensors, datasets, and tasks, and that its benefit is largest out of distribution. The encoder incurs a higher computational cost, which we consider acceptable given the continued growth of GPU throughput and memory, and because each recording is encoded only once while all downstream tasks operate on the stored embeddings. Our contribution is therefore not a superior architecture but a simple recipe for pooling radar IQ data from different sensors into one larger pretraining corpus, since more data yields more general representations. We expect such pooled corpora to enable radar foundation models and cross-sensor synergies that the limited data available per sensor has so far precluded.

{\small
\bibliographystyle{ieeenat_fullname}
\bibliography{egbib}

\begin{thebibliography}{36}
\providecommand{\natexlab}[1]{#1}
\providecommand{\url}[1]{\texttt{#1}}
\expandafter\ifx\csname urlstyle\endcsname\relax
  \providecommand{\doi}[1]{doi: #1}\else
  \providecommand{\doi}{doi: \begingroup \urlstyle{rm}\Url}\fi

\bibitem[Ding et~al.(2022)Ding, Xu, Tang, Xu, Wang, and Tao]{ding2022source}
Ning Ding, Yixing Xu, Yehui Tang, Chao Xu, Yunhe Wang, and Dacheng Tao.
\newblock Source-free domain adaptation via distribution estimation.
\newblock In \emph{Proceedings of the IEEE/CVF conference on computer vision
  and pattern recognition}, pages 7212--7222, 2022.

\bibitem[Engel et~al.(2025)Engel, Mueller, Rendon, Dorschky, Krauss, Ullmann,
  Eskofier, and Vossiek]{10904474}
Lukas Engel, Jonas Mueller, Eduardo Javier~Feria Rendon, Eva Dorschky, Daniel
  Krauss, Ingrid Ullmann, Bjoern~M. Eskofier, and Martin Vossiek.
\newblock Advanced millimeter wave radar-based human pose estimation enabled by
  a deep learning neural network trained with optical motion capture ground
  truth data.
\newblock \emph{IEEE Journal of Microwaves}, 5\penalty0 (2):\penalty0 373--387,
  2025.

\bibitem[Gurbuz et~al.(2020)Gurbuz, Rahman, Kurtoglu, Macks, and
  Fioranelli]{10.1117/12.2559155}
Sevgi~Z. Gurbuz, M.~Mahbubur Rahman, Emre Kurtoglu, Trevor Macks, and Francesco
  Fioranelli.
\newblock {Cross-frequency training with adversarial learning for radar
  micro-Doppler signature classification (Rising Researcher)}.
\newblock In \emph{Radar Sensor Technology XXIV}, page 114080A. International
  Society for Optics and Photonics, SPIE, 2020.

\bibitem[Higgins et~al.(2017)Higgins, Matthey, Pal, Burgess, Glorot, Botvinick,
  Mohamed, and Lerchner]{higgins2017beta}
Irina Higgins, Loic Matthey, Arka Pal, Christopher Burgess, Xavier Glorot,
  Matthew Botvinick, Shakir Mohamed, and Alexander Lerchner.
\newblock beta-vae: Learning basic visual concepts with a constrained
  variational framework.
\newblock In \emph{International conference on learning representations}, 2017.

\bibitem[Ho et~al.(2024)Ho, Cheng, Kuan, Jiang, Chai, Huang, Lin, and
  Hwang]{ho2024rt}
Yuan-Hao Ho, Jen-Hao Cheng, Sheng~Yao Kuan, Zhongyu Jiang, Wenhao Chai,
  Hsiang-Wei Huang, Chih-Lung Lin, and Jenq-Neng Hwang.
\newblock Rt-pose: A 4d radar tensor-based 3d human pose estimation and
  localization benchmark.
\newblock In \emph{European Conference on Computer Vision}, pages 107--125.
  Springer, 2024.

\bibitem[Huang et~al.(2025)Huang, Prabhakara, Chen, Karhade, Ramanan, O'toole,
  and Rowe]{huang2025towards}
Tianshu Huang, Akarsh Prabhakara, Chuhan Chen, Jay Karhade, Deva Ramanan,
  Matthew O'toole, and Anthony Rowe.
\newblock Towards foundational models for single-chip radar.
\newblock In \emph{Proceedings of the IEEE/CVF International Conference on
  Computer Vision}, pages 24655--24665, 2025.

\bibitem[Kingma and Welling(2013)]{kingma2013auto}
Diederik~P Kingma and Max Welling.
\newblock Auto-encoding variational bayes.
\newblock \emph{arXiv preprint arXiv:1312.6114}, 2013.

\bibitem[Krauss et~al.(2024)Krauss, Engel, Ott, Br{\"a}unig, Richer, Gambietz,
  Albrecht, Hille, Ullmann, Braun, et~al.]{krauss2024review}
Daniel Krauss, Lukas Engel, Tabea Ott, Johanna Br{\"a}unig, Robert Richer,
  Markus Gambietz, Nils Albrecht, Eva~M Hille, Ingrid Ullmann, Matthias Braun,
  et~al.
\newblock A review and tutorial on machine learning-enabled radar-based
  biomedical monitoring.
\newblock \emph{IEEE open journal of engineering in medicine and biology},
  5:\penalty0 680--699, 2024.

\bibitem[Kurtoğlu et~al.(2022)Kurtoğlu, Gurbuz, Malaia, Griffin, Crawford,
  and Gurbuz]{9660776}
Emre Kurtoğlu, Ali~C. Gurbuz, Evie~A. Malaia, Darrin Griffin, Chris Crawford,
  and Sevgi~Z. Gurbuz.
\newblock Asl trigger recognition in mixed activity/signing sequences for rf
  sensor-based user interfaces.
\newblock \emph{IEEE Transactions on Human-Machine Systems}, 52\penalty0
  (4):\penalty0 699--712, 2022.

\bibitem[Lee et~al.(2023)Lee, Kini, Peng, Ma, and Hwang]{lee2023hupr}
Shih-Po Lee, Niraj~Prakash Kini, Wen-Hsiao Peng, Ching-Wen Ma, and Jenq-Neng
  Hwang.
\newblock Hupr: A benchmark for human pose estimation using millimeter wave
  radar.
\newblock In \emph{2023 IEEE/CVF Winter Conference on Applications of Computer
  Vision (WACV)}, pages 5704--5713. IEEE, 2023.

\bibitem[Li et~al.(2021)Li, Wang, Yang, and Fu]{9658500}
Xinrong Li, Xiaodong Wang, Qing Yang, and Song Fu.
\newblock Signal processing for tdm mimo fmcw millimeter-wave radar sensors.
\newblock \emph{IEEE Access}, 9:\penalty0 167959--167971, 2021.

\bibitem[Lin et~al.(2026)Lin, Huang, Ding, Zhao, Wang, Wang, and
  Xi]{lin2026active}
Mingzhi Lin, Teng Huang, Han Ding, Cui Zhao, Fei Wang, Ge Wang, and Wei Xi.
\newblock Active domain adaptation for mmwave-based har via r{\'e}nyi
  entropy-based uncertainty estimation.
\newblock \emph{IEEE Transactions on Mobile Computing}, 2026.

\bibitem[Mueller et~al.(2024)Mueller, Engel, Krauss, Ullmann, Vossiek,
  Eskofier, and Dorschky]{mueller2024end}
Jonas Mueller, Lukas Engel, Daniel Krauss, Ingrid Ullmann, Martin Vossiek,
  Bjoern~M Eskofier, and Eva Dorschky.
\newblock End-to-end learning for human pose estimation from raw millimeter
  wave radar data.
\newblock In \emph{2024 58th Asilomar Conference on Signals, Systems, and
  Computers}, pages 1495--1499. IEEE, 2024.

\bibitem[Mueller et~al.(2026{\natexlab{a}})Mueller, Gambietz, Weiss, Krauss,
  and Eskofier]{mueller2026learning}
Jonas~Leo Mueller, Markus Gambietz, Alexander Weiss, Daniel Krauss, and
  Bjoern~M Eskofier.
\newblock Learning biomechanically plausible human motion from sparse radar
  point clouds.
\newblock \emph{arXiv preprint arXiv:2608.03637}, 2026{\natexlab{a}}.

\bibitem[Mueller et~al.(2026{\natexlab{b}})Mueller, Hoefler, Zanca, Jami,
  Altstidl, and Eskofier]{mueller2026you}
Jonas~Leo Mueller, Sebastian Hoefler, Dario Zanca, Naga Venkata Sai~Jitin Jami,
  Thomas Altstidl, and Bjoern~M Eskofier.
\newblock You only flow once: Calibrated and real-time radar pose estimation
  with multi-hypothesis normalizing flows.
\newblock \emph{arXiv preprint arXiv:2608.09579}, 2026{\natexlab{b}}.

\bibitem[M{\"u}ller et~al.(2026)M{\"u}ller, Engel, Dorschky, Krauss, Ullmann,
  Vossiek, and Eskofier]{muller2026radproposer}
Jonas~Leo M{\"u}ller, Lukas Engel, Eva Dorschky, Daniel Krauss, Ingrid Ullmann,
  Martin Vossiek, and Bj{\"o}rn~M Eskofier.
\newblock Radproposer: Probabilistic radar tensor human pose estimation that
  knows its limits.
\newblock In \emph{2026 International Joint Conference on Neural Networks
  (IJCNN)}, pages 1--8. IEEE, 2026.

\bibitem[Pinyoanuntapong et~al.(2023)Pinyoanuntapong, Ali, Jakkala, Wang, Lee,
  Peng, Chen, and Sun]{pinyoanuntapong2023gaitsada}
Ekkasit Pinyoanuntapong, Ayman Ali, Kalvik Jakkala, Pu Wang, Minwoo Lee,
  Qucheng Peng, Chen Chen, and Zhi Sun.
\newblock Gaitsada: Self-aligned domain adaptation for mmwave gait recognition.
\newblock In \emph{2023 IEEE 20th International Conference on Mobile Ad Hoc and
  Smart Systems (MASS)}, pages 218--226. IEEE, 2023.

\bibitem[Rahman and Gurbuz(2023)]{rahman2023self}
Mohammad~Mahbubur Rahman and Sevgi~Zubeyde Gurbuz.
\newblock Self-supervised contrastive learning for radar-based human activity
  recognition.
\newblock In \emph{2023 IEEE Radar Conference (RadarConf23)}, pages 1--6. IEEE,
  2023.

\bibitem[Richards(2005)]{richards2005fundamentals}
M.A. Richards.
\newblock \emph{Fundamentals of Radar Signal Processing}.
\newblock Mcgraw-hill, 2005.

\bibitem[Seyfio{\u{g}}lu and G{\"u}rb{\"u}z(2017)]{seyfiouglu2017deep}
Mehmet~Sayg{\i}n Seyfio{\u{g}}lu and Sevgi~Z{\"u}beyde G{\"u}rb{\"u}z.
\newblock Deep neural network initialization methods for micro-doppler
  classification with low training sample support.
\newblock \emph{IEEE Geoscience and Remote Sensing Letters}, 14\penalty0
  (12):\penalty0 2462--2466, 2017.

\bibitem[Seyfio{\u{g}}lu et~al.(2018)Seyfio{\u{g}}lu, {\"O}zbayo{\u{g}}lu, and
  G{\"u}rb{\"u}z]{seyfiouglu2018deep}
Mehmet~Sayg{\i}n Seyfio{\u{g}}lu, Ahmet~Murat {\"O}zbayo{\u{g}}lu, and
  Sevgi~Zubeyde G{\"u}rb{\"u}z.
\newblock Deep convolutional autoencoder for radar-based classification of
  similar aided and unaided human activities.
\newblock \emph{IEEE Transactions on Aerospace and Electronic Systems},
  54\penalty0 (4):\penalty0 1709--1723, 2018.

\bibitem[Seyfioglu et~al.(2019)Seyfioglu, Erol, Gurbuz, and Amin]{8572732}
Mehmet~Saygin Seyfioglu, Baris Erol, Sevgi~Zubeyde Gurbuz, and Moeness~G. Amin.
\newblock Dnn transfer learning from diversified micro-doppler for motion
  classification.
\newblock \emph{IEEE Transactions on Aerospace and Electronic Systems},
  55\penalty0 (5):\penalty0 2164--2180, 2019.

\bibitem[Sheng et~al.(2022)Sheng, Xu, Zhang, and Wang]{sheng2022facilitating}
Zhiyao Sheng, Huatao Xu, Qian Zhang, and Dong Wang.
\newblock Facilitating radar-based gesture recognition with self-supervised
  learning.
\newblock In \emph{2022 19th Annual IEEE International Conference on Sensing,
  Communication, and Networking (SECON)}, pages 154--162. IEEE, 2022.

\bibitem[Shrestha et~al.(2019)Shrestha, Murphy, Johnson, Anbulselvam,
  Fioranelli, Le~Kernec, and Gurbuz]{shrestha2019cross}
Aman Shrestha, Corben Murphy, Ivan Johnson, Ajaymehul Anbulselvam, Francesco
  Fioranelli, Julien Le~Kernec, and Sevgi~Zubeyde Gurbuz.
\newblock Cross-frequency classification of indoor activities with dnn transfer
  learning.
\newblock In \emph{2019 IEEE Radar Conference (RadarConf)}, pages 1--6. IEEE,
  2019.

\bibitem[Stadelmayer et~al.(2021)Stadelmayer, Santra, Weigel, and
  Lurz]{9464267}
Thomas Stadelmayer, Avik Santra, Robert Weigel, and Fabian Lurz.
\newblock Data-driven radar processing using a parametric convolutional neural
  network for human activity classification.
\newblock \emph{IEEE Sensors Journal}, 21\penalty0 (17):\penalty0 19529--19540,
  2021.

\bibitem[Sun et~al.(2020)Sun, Petropulu, and Poor]{sun2020mimo}
Shunqiao Sun, Athina~P Petropulu, and H~Vincent Poor.
\newblock Mimo radar for advanced driver-assistance systems and autonomous
  driving: Advantages and challenges.
\newblock \emph{IEEE Signal Processing Magazine}, 37\penalty0 (4):\penalty0
  98--117, 2020.

\bibitem[Tanaka et~al.(2025)Tanaka, Ikeda, and
  Yonemoto]{tanaka2025comprehensive}
Tomoya Tanaka, Tomonori Ikeda, and Ryo Yonemoto.
\newblock Comprehensive deployment-oriented assessment for cross-environment
  generalization in deep learning-based mmwave radar sensing.
\newblock \emph{arXiv preprint arXiv:2512.13018}, 2025.

\bibitem[Ullmann et~al.(2023{\natexlab{a}})Ullmann, Guendel, Kruse, Fioranelli,
  and Yarovoy]{ullmann2023radar}
Ingrid Ullmann, Ronny~G Guendel, Nicolas~Christian Kruse, Francesco Fioranelli,
  and Alexander Yarovoy.
\newblock Radar-based continuous human activity recognition with multi-label
  classification.
\newblock In \emph{2023 IEEE SENSORS}, pages 1--4. IEEE, 2023{\natexlab{a}}.

\bibitem[Ullmann et~al.(2023{\natexlab{b}})Ullmann, Guendel, Kruse, Fioranelli,
  and Yarovoy]{ullmann2023survey}
Ingrid Ullmann, Ronny~G Guendel, Nicolas~Christian Kruse, Francesco Fioranelli,
  and Alexander Yarovoy.
\newblock A survey on radar-based continuous human activity recognition.
\newblock \emph{IEEE Journal of Microwaves}, 3\penalty0 (3):\penalty0 938--950,
  2023{\natexlab{b}}.

\bibitem[Ullmann et~al.(2024)Ullmann, Guendel, Kruse, Fioranelli, and
  Yarovoy]{ullmann2024classification}
Ingrid Ullmann, Ronny~G Guendel, Nicolas~Christian Kruse, Francesco Fioranelli,
  and Alexander Yarovoy.
\newblock Classification strategies for radar-based continuous human activity
  recognition with multiple inputs and multilabel output.
\newblock \emph{IEEE Sensors Journal}, 24\penalty0 (24):\penalty0 40251--40261,
  2024.

\bibitem[Wang et~al.(2021)Wang, Sun, Cheng, Jiang, Deng, Zhao, Liu, Mu, Tan,
  Wang, Liu, and Xiao]{9052469}
Jingdong Wang, Ke Sun, Tianheng Cheng, Borui Jiang, Chaorui Deng, Yang Zhao,
  Dong Liu, Yadong Mu, Mingkui Tan, Xinggang Wang, Wenyu Liu, and Bin Xiao.
\newblock Deep high-resolution representation learning for visual recognition.
\newblock \emph{IEEE Transactions on Pattern Analysis and Machine
  Intelligence}, 43\penalty0 (10):\penalty0 3349--3364, 2021.

\bibitem[Wang et~al.(2026)Wang, Wang, and Wang]{Wang_2026_CVPR}
Shengpeng Wang, Kuangyu Wang, and Wei Wang.
\newblock Rauf: Learning the spatial uncertainty field of radar.
\newblock In \emph{Proceedings of the IEEE/CVF Conference on Computer Vision
  and Pattern Recognition (CVPR)}, pages 42474--42483, 2026.

\bibitem[Wei et~al.(2026)Wei, Fang, Chetty, Cho, and
  Bianchi-Berthouze]{wei2026maepose}
Xijia Wei, Yuan Fang, Kevin Chetty, Youngjun Cho, and Nadia Bianchi-Berthouze.
\newblock Maepose: Self-supervised spatiotemporal learning for human pose
  estimation on mmwave video.
\newblock \emph{arXiv preprint arXiv:2605.00242}, 2026.

\bibitem[Wu et~al.(2026)Wu, Fioranelli, and Gao]{11316534}
Yizhuo Wu, Francesco Fioranelli, and Chang Gao.
\newblock Radmamba: Efficient human activity recognition through a radar-based
  micro-doppler-oriented mamba state-space model.
\newblock \emph{IEEE Transactions on Radar Systems}, 4:\penalty0 261--272,
  2026.

\bibitem[Xue et~al.(2023)Xue, Cao, Miao, Ju, Hu, Zhang, and Su]{xue2023towards}
Hongfei Xue, Qiming Cao, Chenglin Miao, Yan Ju, Haochen Hu, Aidong Zhang, and
  Lu Su.
\newblock Towards generalized mmwave-based human pose estimation through signal
  augmentation.
\newblock In \emph{Proceedings of the 29th Annual International Conference on
  Mobile Computing and Networking}, pages 1--15, 2023.

\bibitem[Zhao et~al.(2023)Zhao, Lu, Wang, Trigoni, and
  Markham]{zhao2023cubelearn}
Peijun Zhao, Chris~Xiaoxuan Lu, Bing Wang, Niki Trigoni, and Andrew Markham.
\newblock Cubelearn: End-to-end learning for human motion recognition from raw
  mmwave radar signals.
\newblock \emph{IEEE Internet of Things Journal}, 10\penalty0 (12):\penalty0
  10236--10249, 2023.

\end{thebibliography}
}

\end{document}